\documentclass{article}
\usepackage{ijcai17}

\usepackage{times}
\usepackage{url}
\usepackage{amsmath}
\usepackage{latexsym}
\usepackage{caption}
\usepackage{amssymb}
\usepackage{booktabs, multicol, multirow}
\usepackage{float}
\usepackage{url}
\usepackage{graphicx}
\usepackage{pdfpages}
\usepackage[draft]{hyperref}

\title{Learning Sentence Representation with Guidance of Human Attention}
\author{Shaonan Wang, Jiajun Zhang and Chengqing Zong\\
University of Chinese Academy of Sciences, Beijing, China\\
National Laboratory of Pattern Recognition, CASIA, Beijing, China\\
\{shaonan.wang,jjzhang,cqzong\}@nlpr.ia.ac.cn\\
}
\author{Shaonan Wang$^{1,2}$, Jiajun Zhang$^{1,2}$, Chengqing Zong$^{1,2,3}$ \\
  $^1$ National Laboratory of Pattern Recognition, CASIA, Beijing, China \\ $^2$ University of Chinese Academy of Sciences, Beijing, China \\
  $^3$ CAS Center for Excellence in Brain Science and Intelligence Technology, Shanghai, China  \\
  {\tt \{shaonan.wang,jjzhang,cqzong\}@nlpr.ia.ac.cn} \\}

\begin{document}

\maketitle

\begin{abstract}
Recently, much progress has been made in learning general-purpose sentence representations that can be used across domains. However, most of the existing models typically treat each word in a sentence equally. In contrast, extensive studies have proven that human read sentences efficiently by making a sequence of fixation and saccades. This motivates us to improve sentence representations by assigning different weights to the vectors of the component words, which can be treated as an attention mechanism on single sentences. To that end, we propose two novel attention models, in which the attention weights are derived using significant predictors of human reading time, i.e., Surprisal, POS tags and CCG supertags. The extensive experiments demonstrate that the proposed methods significantly improve upon the state-of-the-art sentence representation models.
\end{abstract}

\section{Introduction}

To understand the meaning of a sentence is a prerequisite to solve many linguistic and non-linguistic problems: answer a question, translate the text into another language and so on. Obviously, this requires a good representation of the meaning of a sentence. Recently, neural network based sentence representation models have shown advantages in learning general-purpose sentence embeddings \cite{le2014distributed,kiros2015skip,wieting2016towards}. However, these models typically treat each word in a sentence equally. This is inconsistent with the way that human read and understand sentences, namely reading some words superficially and paying more attention to others. All these factors motivate us to build sentence representation models that can selectively focus on important words, which can be treated as a task-independent attention mechanism. 
	
The main difficulty of introducing the above attention mechanism to a single sentence is the lack of extra information to guide the computation of attention weight. In this paper, we hypothesize that the significant predictors of human reading time are such useful information. So far, extensive studies have proven that word attributes, as represented by POS tag, length, frequency, Surprisal, etc., are all correlated with human reading time \cite{demberg2008data,barrett2016weakly}. This paper focuses on two kinds of predictors: Surprisal as a continuous variable; POS tags and Combinatory Categorial Grammar (CCG) supertags which are discretely variables. Surprisal, proposed by \cite{hale2001probabilistic} and \cite{levy2008expectation}, measures the amount of information conveyed by a particular event. Generally, the higher surprisal value corresponds to the higher processing complexity and more reading time. Moreover, psycholinguistic experiments have shown that readers are more likely to fixate on words from open syntactic categories (verbs, nouns, adjectives) than on closed category items like prepositions and conjunctions \cite{rayner1998eye}. These findings indicate that the above factors are crucial for simulating human attention in reading.

In this paper, we propose two novel attention approaches which are called attention model with Surprisal (ATT-SUR) and attention model with POS tag or CCG supertag (ATT-POS/ATT-CCG), respectively, to improve sentence representations. One approach utilizes Surprisal directly as the attention weight. The other approach builds attention model with the help of POS tag and CCG supertag vectors which are trained together with word embeddings. Aiming at enhancing semantic representation of sentences, the proposed attention models are then combined with two state-of-the-art (unsupervised/semi-supervised) sentence representation models. Furthermore, we perform extensive quantitative and qualitative analysis to shed light on the principle of the proposed attention models and its relation with human attention mechanism in reading.

To summarize, our main contributions include:

\begin{itemize}
\item We present two simple but efficient attention models for sentence representations, which can also be seen as a general framework of integrating predictors of human reading time into sentence representation models. 
\item We have evaluated our approaches on 24 SemEval datasets on semantic textual similarity (STS) tasks, which contain a  wide range of domains. The results show that our approaches can significantly improve semantic representation of sentences.
\item Experimental results have indicated that the proposed attention models can selectively focus on important words and successfully predict human reading time. 

\end{itemize}

\section{Background}
We introduce two main approaches for learning general-purpose sentence representations according to the training material used: unsupervised methods trained on raw text corpora (Section 2.1), and semi-supervised methods trained on out-of-domain annotated text corpora(Section 2.2). For each method, we \textit{choose the state-of-the-art model as our baseline} to incorporate the proposed attention models.

\subsection{Unsupervised Methods}
\cite{mikolov2013efficient} constructed a learning criterion for obtaining word representations from unlabeled data, by predicting a word from its surrounding words. Afterwards, several approaches for learning sentence representations were proposed, extending this strategy at the sentence level by predicting a sentence from its adjacent sentences \cite{kiros2015skip,hill2016learning,kenter2016siamese}, or by learning extra sentence embeddings in the learning process of word embeddings \cite{le2014distributed,wang2016cse}. 

Among them, the Siamese CBOW (SCBOW) model introduced by  \cite{kenter2016siamese} is the best performing method on multiple test sets. This method utilizes successive sentences as the training corpus, e.g., sentences in the article, and trains with categorical cross-entropy method. We briefly describe the SCBOW model below:

\textbf{Baseline 1:} The SCBOW model represents sentences by averaging  the embeddings of its constituent words. Given a word sequence with length $n$: $x=<x_1, x_2, ..., x_n>$, the sentence representation model is described as:

\begin{equation}
g_{sentence}(x)=\frac{1}{n}\sum_{i=1}^{n}W_{w}^{x_i},
\end{equation}
where $W_w$ is the word embedding matrix. For a pair of sentences $(s_i, s_j)$, we define the set $S^+$ as the sentences that occur next to the sentence $s_i$, and $S^-$, a set of randomly chosen sentences which are not in $S^+$. The probability $p_\theta(s_i, s_j)$ reflects how likely the sentence pairs are adjacent to each other in the training data and is computed as:

\begin{equation}
p_\theta(s_i, s_j)=\frac{exp(cos(s_i^\theta ,s_j^\theta ))}{\sum_{s_k \in{\lbrace S^+\cup S^- \rbrace}}exp(cos(s_i^\theta ,s_k^\theta ))},
\end{equation}
where $s_x^\theta$ denotes the embedding of sentence $s_x$. The objective  function is defined as:
 
\begin{equation}
L = - \sum_{s_j \in{\lbrace S^+\cup S^- \rbrace}} p(s_i, s_j) \cdot log(p_{\theta}(s_i,s_j)),
\end{equation}
where $p(s_i, s_j)$ is the target probability the network should produce, which is  $\frac{1}{|S^+|}$ if $s_j\in S^+$ and $0$ if $s_j\in S^-$.

\subsection{Semi-Supervised Methods}
Lately, various models for learning distributed sentence representations have been proposed, ranging from simple additional composition of the word vectors to sophisticated architectures such as convolution neural networks and recurrent neural networks. However, sentence representations, generated by most of the existing work, are tuned only for their respective task. More recently, \cite{wieting2016towards} proposed the Paragram-Phrase (PP) model, which learns general-purpose sentence embeddings with supervision from the Paraphrase Database (PPDB) \cite{ganitkevitch2013ppdb} . This simple method is extremely efficient, outperforming more complex models (e.g., LSTM model), and even competitive with systems tuned for particular tasks.

\textbf{Baseline 2:} The PP model constructs sentence representations with the word averaging model defined in equation (1). The training data consists of a set of phrase pairs ($x_1$, $x_2$) from the PPDB dataset and negative examples ($t_1$, $t_2$) which are the most similar phrase pairs to ($x_1$, $x_2$) generated in a mini-batch during optimization. The PP model uses a max-margin objective function to train sentence embeddings by maximizing the distance between positive examples and negative examples:

\begin{equation} \small
\begin{array}{l}
_{{W_w}}^{\min }\frac{1}{{\left| X \right|}}(\sum\limits_{({x_1},{x_2}) \in X} {\max (0,1 - W_w^{{x_1}} \cdot W_w^{{x_2}} + }W_w^{{x_1}} \cdot W_w^{{t_1}}) + \\
 \max (0,1 - W_w^{{x_1}} \cdot W_w^{{x_2}} + W_w^{{x_2}} \cdot W_w^{{t_2}})) + \lambda {\left\| {{W_{{w_{i}}}} - {W_w}} \right\|^2}
\end{array}
\end{equation}
Where $\lambda$ is the regularization parameter, $\left| X \right|$ is the length of training paraphrase pairs, $W_w$ is the current word vector matrix, and $W_{w_{i}}$ is the initial word vector matrix.

\section{Attention-based Sentence Representation Model}
This section introduces the proposed attention models (Section 3.1), and how to integrate them into sentence representation models (Section 3.2).

\subsection{Attention Models}
\subsubsection{ATT-SUR: Attention Model with Surprisal.}
Surprisal, also known as self-information, measures the amount of information conveyed by the target. In language processing, it is defined as:

\begin{equation}
s^{x_t}=-log(P(x_t|x_1,...,x_{t-1})),
\end{equation}
where the Surprisal $s^{x_t}$ corresponds to the negative logarithm of the conditional probability of word $x_t$ given the sentential context $x_1,..., x_{t-1}$. 

Based on the assumption that words with higher surprisal value convey more information and should gain more attention, we directly use the value of Surprisal as the attention weight. The proposed ATT-SUR model is computed as:

\begin{equation}
attention(x_t)=\frac{exp(s^{x_t})}{\sum_{i\in[1,...,n]}exp(s^{x_i})},
\end{equation}
where $n$ is the length of the word sequence.

\subsubsection{ATT-POS (ATT-CCG): Attention Model with POS Tags (CCG Supertags).}
In this work we hypothesize that POS tag and CCG supertag of words are useful factors in building the attention model for sentence representations. For instance, given a sentence \texttt{a\#DT man\#NN with\#IN a\#DT hard\#JJ hat\#NN is\#VBZ dancing\#VBG}, the optimal sentence representation should give more weights to words with \texttt{NN}, \texttt{JJ}, \texttt{VBZ} and \texttt{VBG} tags, and less weights to words with \texttt{DT} and \texttt{IN} tags. To model the above observation, we assign a vector to each POS tag (CCG supertag) and compute the dot product with the corresponding word embedding vectors. The result is a scalar parameter which determines the relative power of each of the POS tag (CCG supertag), which is described as:

\begin{equation}
attention(x_t)=\frac{exp(W_w^{x_t} \cdot{W_c^{x_t}) }}{\sum_{i\in[1,..,n]}exp(W_w^{x_t} \cdot{W_c^{x_t}}) },
\end{equation}
where $W_w \in \Re^{V_w \times d} $ is the word embedding matrix and $W_c \in \Re^{V_c \times d}$ is the word class matrix with each line represents a POS tag (CCG tag) vector.

\subsection{Incorporating the Attention Models into Sentence Representations}

To incorporate attention mechanism into sentence representations learned by the SCBOW model and the PP model, we use weighted summation of word embeddings instead of the averaging model in equation (1), and the weight is calculated by equation (6) or (7). The attention-based sentence representation model is computed as:

\begin{equation}
g_{sentence}(x)=\frac{1}{n}\sum_{i=1}^{n}attention(x_i)W_{w}^{x_i}
\end{equation}

\section{Experiments and Results}
\subsection{Datasets}
Following the parameter settings in \cite{kenter2016siamese} and \cite{wieting2016towards}, the SCBOW model uses the Toronto Book Corpus\footnote{The corpus can be downloaded from \url{http://www.cs.toronto.edu/~mbweb/}.} which contains 7,087 books collected from the web. The PP model is first trained with the smaller PPDB dataset (version XL) for 10 epochs and then trained for another 10 epochs on a much larger PPDB dataset (version XXL). To evaluate the performance of our models, we use 24 datasets from STS task, covering a wide range of domains like news, image and video descriptions, glosses, twitter, machine translation evaluation and so on.

\subsection{Experimental Settings}
In this paper, we use the Stanford POS tagger\footnote{ \url{http://nlp.stanford.edu/software/tagger.shtml}} and the C\&C tool\footnote{\url{http://svn.ask.it.usyd.edu.au/trac/candc/wiki/Download}} to assign POS tags and CCG supertags, respectively, for words in the training and testing datasets. In all the models, we randomly initialize the POS tag and CCG supertag vectors with 300-dimension vectors, by drawing from a normal distribution with $\mu=0.0$ and $\sigma=0.01$. In the ATT-SUR model, the Surprisal is calculated by a state-of-the-art large-scale neural language model released by \cite{jozefowicz2016exploring}. Moreover, we also train a 5 order n-gram language model with modified Kneser-Ney smoothing, but the performance is slightly worse. Hence we only report the results of the neural language model. In the experiment, we set surprisal value $x$ as  $min(max(0, x), 10)$.

In the SCBOW model, we use two negative examples and initialize the embedding layer with the pre-trained word embeddings\footnote{The word embeddings is available at \url{https://github.com/mmihaltz/word2vec-GoogleNews-vectors}}. Then we train the model using AdaDelta for one epoch with the initial learning rate of 0.001 and the batch size of 100. For the PP model, phrases in the training dataset are not sentences or even constituents, causing worse tagging results. Hence we use the SICK data set, which consists of 10,000 English sentence pairs with human annotation, to train the attention models after training the PP model\footnote{The trained model is available at \url{http://ttic.uchicago.edu/~wieting/}}. The attention models are trained by AdaGrad for ten epochs with  initial learning rate of 0.05. \textit{The code for training and evaluation will be released.}

\subsection{Textual Similarity}
Following previous work in evaluating the STS task, we use the Pearson's correlation coefficient to examine relationships between the human ratings and the predicted cosine similarity scores of sentence pairs. Table 1 displays the results of the two baseline models (PP-Base, SCBOW-Base), TF-IDF baseline model\footnote{This model represents sentence meaning by the same method with ATT-SUR model and replaces value of word suprisal with \texttt{tf-idf}, in which the \texttt{tf-idf} is computed with all the textual similarity datasets and each sentence is viewed as a \texttt{document}.} and the proposed attention-based sentence representation models on all 24 textual similarity datasets\footnote{We have also conducted experiments on baselines like simply reserve nouns, verbs, adjectives and adverbs. The average results on STS datasets are between “Base” and “TF-IDF” model with some results better or poorer than “Base” or “TF-IDF” on specific datasets.}. In addition, we also include three more baselines: the SkipThought (ST) \cite{kiros2015skip} model, that utlizes a encoder-decoder model and produces highly generic sentence representations\footnote{We employ the trained model relased in \url{https://github.com/ryankiros/skip-thoughts} for evaluation.}. The ParaphraseVec (PV) \cite{le2014distributed} model, as an extension of the CBOW and Skip-gram models \cite{mikolov2013efficient}, represents sentences as fixed-length vectors effeciently in a non-compositional way. The DictRep (DictR) \cite{hill2016dict} model, which learns sentence representations by mapping dictionary definitions to a pre-trained word embeddings of the words defined by those definitions, achieves the best performance on STS task within eight sentence representation models \cite{hill2016learning}. 

\newcommand{\tabincell}[2]{\begin{tabular}{@{}#1@{}}#2\end{tabular}}
\begin{table*}[htb] \scriptsize
\centering
\caption{Pearson rank correlation of model predictions with subject similarity ratings on SemEval textual similarity datasets. The bold scores in each row are the best result in the Baselines column, PP model column and the SCBOW model column, respectively. \texttt{Base} denotes the model without attention mechanism. Results of PV and DictR are reprinted from \protect\cite{hill2016learning} where \textit{they are reported for only Semeval 2014 datasets} in two-digit precision.}

\begin{tabular}{c||ccc||ccccc||ccccc}
\toprule
  &  \multicolumn{3}{c||}{Baselines}&  \multicolumn{5}{c||}{PP}& \multicolumn{5}{c}{SCBOW} \\
 \cline{2-14}&ST&PV&DictR& Base & TF-IDF & ATT-SUR & ATT-POS & ATT-CCG & Base & TF-IDF & ATT-SUR & ATT-POS & ATT-CCG\\
\hline
 MSRpar& 0.168&- & -&	0.476& 0.465 &	0.486& 0.497& \textbf{0.499}&	0.429& 0.412&	\textbf{0.437}&	0.414&	0.419\\
 MSRvid& 0.437&-& -&	0.774&	0.792 & 0.802&	\textbf{0.846}&0.842&0.620&	0.611&0.672&	0.702&	\textbf{0.734}\\
OnWN& 0.413&- &- &	0.714&	0.725& 0.726&	0.725&\textbf{0.727}&0.687&	0.688&0.677&	0.695&	\textbf{0.696}\\
SMTeurop&0.413&-& -&	0.481&	\textbf{0.521}&0.505&	0.493&0.493&0.537&	0.542&0.533&	0.538&	\textbf{0.552}\\
SMTnews&0.335&-&- &	0.652&	0.658&0.663&	0.664&\textbf{0.666}&0.523&	0.525&0.544&	0.541&	\textbf{0.557}\\
\midrule
2012 Average&0.353&-& -&	0.619& 0.632&	0.636&	\textbf{0.645}& \textbf{0.645} &0.559&0.556 &	0.573&	0.578&	\textbf{0.592}\\
\midrule
FNWN&0.201&-& -&	0.476&	0.500&0.502&	0.498&\textbf{0.507}&0.378&	0.375&0.350&	0.383&	\textbf{0.392}\\
OnWN& 0.241&-& -&	0.738&	0.745&0.760&	0.779&\textbf{0.793}&0.584&	0.585&\textbf{0.649}&	0.609&	0.583\\
headlines& 0.334&-& -& 0.733&0.738&\textbf{0.748}& 0.737&0.736&	0.693&	0.688&0.705&	0.704&	\textbf{0.711}\\
\midrule
2013 Average&0.259&-&-&	0.649&	0.661&0.67&	0.671&\textbf{0.679}&0.552&	0.549&\textbf{0.568}&	0.565&	0.562\\
\midrule
OnWN&  0.354&0.51& \textbf{0.85}&	0.812&	0.810&0.824& 0.836&\textbf{0.841}&	0.686&	0.682&\textbf{0.729}&	0.705&	0.693\\
deft-forum& 0.249&0.33&\textbf{0.49}&	0.540&	0.544&0.554&	\textbf{0.558}&0.552&0.400&  0.400&0.409&	0.409&	\textbf{0.421}\\
deft-news& 0.369&0.42&\textbf{0.65}&	0.739&	0.738&0.746&	0.741& \textbf{0.755}&0.724& 0.719&0.715&	0.726&	\textbf{0.733}\\
headlines& 0.332&0.46&\textbf{0.57}&	0.707&	0.713&0.721&	\textbf{0.723} &0.722&0.652& 0.653&0.657&	0.666&	\textbf{0.668}\\
images& 0.375&0.32&\textbf{0.71}&	0.805&	0.808&0.808&	\textbf{0.831}& \textbf{0.831}&0.660&	0.656&0.682&	0.733&	\textbf{0.761}\\
tweets& 0.399&0.54&\textbf{0.67}&	0.769&	0.773&0.777&	\textbf{0.791}&0.777&0.708&	0.699&0.713&	\textbf{0.754}&	0.723\\
\midrule
2014 Average&0.346 &0.43&\textbf{0.67}&	0.729&	0.731&0.738&	\textbf{0.746}& \textbf{0.746} &0.638& 0.634&	0.651&	0.665&	\textbf{0.666}\\
\midrule
ans-forums& 0.299&-&-&	0.691&	0.684&\textbf{0.694}&	0.690&0.692&0.476&0.475&	\textbf{0.531}&	0.491&	0.512\\
ans-students& 0.361&-&-&	0.781&	\textbf{0.785}&0.782&	0.791& 0.783 &0.723&0.724&	0.722&	0.715&	\textbf{0.732}\\
belief&0.387&-&-&0.773&	0.782&0.783&	0.780& \textbf{0.784}&0.584& 0.594&0.598&	0.600&	\textbf{0.603}\\
headlines& 0.415&-&-&	0.764&	0.768&0.773&	0.773& \textbf{0.774}&0.713& 0.715&0.723&	0.720&	\textbf{0.727}\\
images& 0.423&-&-&	0.837&	0.838&0.841&	0.851& \textbf{0.853}& 0.740&0.734&0.740&	0.773&	\textbf{0.787}\\
\midrule
2015 Average& 0.377&-&-&	0.769&	0.771&0.775&	\textbf{0.777}& \textbf{0.777}&0.647&0.648&	0.663&	0.660&	\textbf{0.672}\\
\midrule
answer& 0.439&-&-&	\textbf{0.670}&	0.659&0.669&	0.660&0.645&\textbf{0.398}&	0.396&0.385&	0.379&	0.391\\
deadlines& 0.370&-&-&	0.699&	0.707&\textbf{0.712}&	0.709&0.701&0.683&	0.668&0.696&	\textbf{0.705}&	0.701\\
plagiarism& 0.559&-&-&	0.802&	0.815&0.819&	0.814& \textbf{0.825}&0.708& 0.699&0.731&	\textbf{0.736}&	\textbf{0.736}\\
postediting& 0.690&-&-&	0.828&	0.830&\textbf{0.831}&	0.824&0.830&\textbf{0.708}&	0.697&0.713&	0.700&	0.706\\
question&0.242&-&-&	0.535&0.551	&0.561&	0.582& \textbf{0.585}&0.580&  0.475&	\textbf{0.672}&	0.623&	0.652\\
\midrule
2016 Average& 0.46&-&-&	0.707&	0.712&\textbf{0.718}&	0.717& \textbf{0.718}&0.615& 0.587&	\textbf{0.639}&	0.629&	0.637\\
\bottomrule
\end{tabular}
\end{table*}        

Comparing the results of the PP-Base, SCBOW-Base and three baselines models, we can see that the PP-Base and DictR models outperform other models by a large margin on most datasets. The PP model is trained with PPDB, which is carefully constructed from a big bilingual parallel corpus. The DictR model utilizes the  corpora of dictionary definitions from Wordnet, Wiktionary and online dictionaries. Clearly, these structured corpora consist of rich semantic information which is extremely helpful for representing sentence meaning. These results suggest that further work of learning general-purpose sentence representations should consider structured corpora, which are widely available on the web.

As it is shown in Table 1, our attention-based sentence representation models outperform state-of-the-art approaches on almost all datasets. However, there are three exceptions: MSRpar-2012, answer-2016 and postediting-2016. Going back to the testing datasets, we find that sentences in MSRpar-2012 and postediting-2016 are really long, with an average length of 20.63 and 19.36. The answer-2016 dataset contains large content of daily conversation, in which function words are in majority. These characteristics may be the reason why our models do not fit in. Another observation is that the performance of the ATT-CCG model is better than the ATT-SUR model and the ATT-POS model, which indicates that the word class with more fine-grained categories (e.g., CCG supertag) is more suitable for building the attention model to improve sentence representations. 

\subsection{Effects of the Attention Models}
To delve deeper into how our models work, we examine the attention weights calculated by the three different attention models and then investigate what have POS tag and CCG supertag vectors learned.

\subsubsection{Do ATT-SUR, ATT-POS and ATT-CCG Differ?}

The ATT-SUR model uses Surprisal directly as the word attention weight, while the ATT-POS model and ATT-CCG model train the POS tag vectors and CCG supertag vectors, respectively, and calculate the attention weight by dot product with the corresponding word embeddings. Due to different mechanism, we expect to see some difference in the learned attention values. Table 2 shows the lowest (highest) attention values of each word by averaging their attention values in all test datasets.

\begin{table*}\footnotesize
\centering
\caption{The five words with the lowest (highest) attention values generated by the ATT-SUR, ATT-POS and ATT-CCG models.}
\begin{tabular}{cccccccccc}
\toprule
\multicolumn{2}{c}{ATT-SUR}&\multicolumn{2}{c}{ATT-POS\tiny{(SCBOW)}}&\multicolumn{2}{c}{ATT-CCG\tiny{(SCBOW)}}&\multicolumn{2}{c}{ATT-POS\tiny{(PP)}}&\multicolumn{2}{c}{ATT-CCG\tiny{(PP)}}\\
\midrule
low&high&low&high&low&high&low&high&low&high\\
\midrule
versa&stenographers&to&skillet&and&skillet&who&revolve&anthropology&bienging\\
cons&dukes&and&uterus&to&colander&consists&centum&vertification&revolve\\
laundering&unsparing&below&colander&a&windowsill&appears&texans&double-check&firstly\\
pong&mason&incredibly&suction&nor&robe&which&depleted&dashes&centum\\
ounce&ringers&really&foal&the&porch&seems&armenia&dash&depleted\\
\bottomrule
\end{tabular}
\end{table*}

From Table 2, as expected, we observe quite different patterns within different attention models. For the ATT-POS\tiny{(SCBOW)}\normalsize \ and ATT-CCG\tiny{(SCBOW)} \normalsize models, five lowest attention words are function words like prepositions and conjunctions, while the highest attention words are content words. However, for the ATT-SUR model, the words with lowest or highest attention value do not show difference in word syntactic category. The five lowest attention words are those often co-occur with a fixed previous word. For example, \texttt{versa} often appears as \texttt{vice versa} and \texttt{laundering} is frequently used in \texttt{money laundering}. The five highest attention words are infrequent words or words occurring in the context that are not included in the training corpus. Moreover, the attention models integrated in the PP model are trained on a much smaller sentence similarity data set. Hence the results reflect corpus-specific characteristics which are dissimilar from the attention models trained in unsupervised setting.

\subsubsection{What do POS Tag Vectors and CCG Supertag Vectors Capture?}

To intuitively show the variations of attention weights by different POS tags\footnote{Here we take POS tag for example and the same conclusion is achieved in CCG supertag.}, we average the attention values of all words within the same tag and show the results in Figure 1.

\begin{figure}[htb]
\centering
\includegraphics[width=80mm,height=25mm]{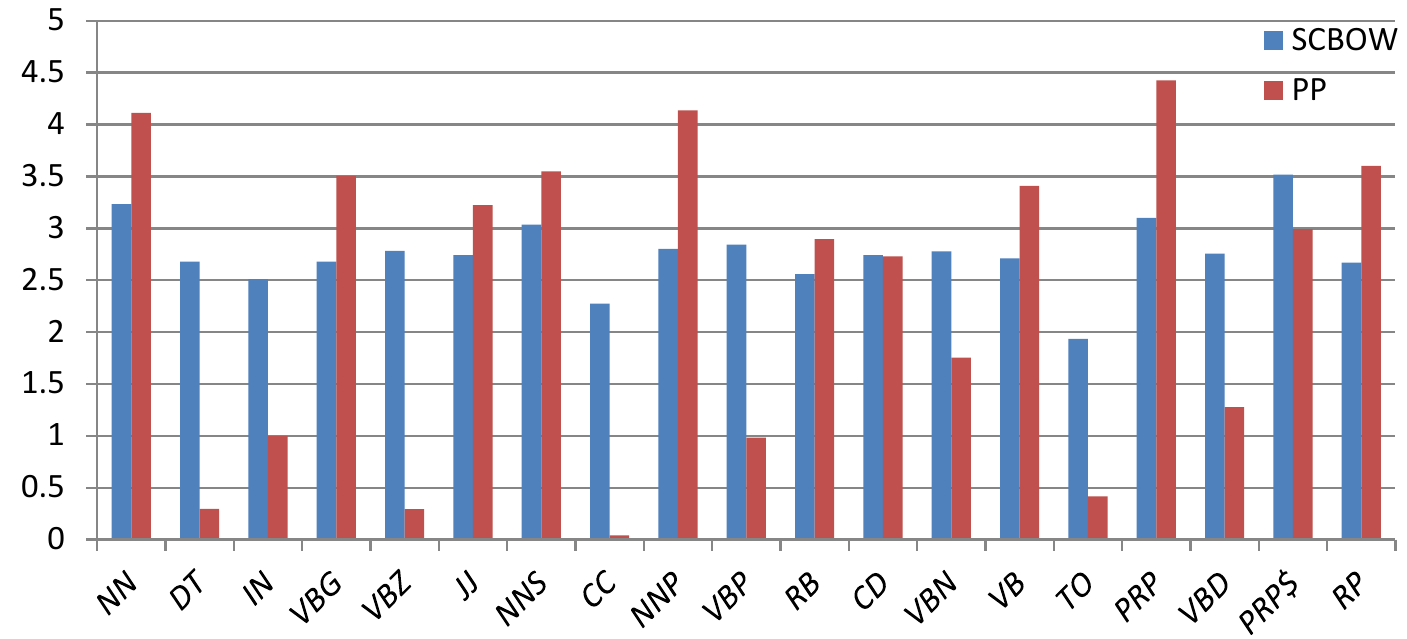}
\caption{Percentage of attention value for the 20 most frequent POS tags, in which \texttt{NN} is the tag with the highest frequency in a collection of test data sets.}
\end{figure}

As it is shown in the figure, POS tag vectors learned in the PP model express more variations in attention values. Nevertheless, the similar trend is achieved for tag vectors learned in both SCBOW and PP models, i.e., nouns, verbs and pronouns get more attention, while prepositions and conjunctions gain less attention. The POS tag vectors and CCG supertag vectors are trained together with word vectors: the attention score is computed as the dot product of a tag vector with the corresponding word vector. Therefore, both $L_2$ norm and direction of a tag vector can affect the results of attention models.

Considering the computation methodology of tag vectors, they might be a mixture of a class of word vectors.  Therefore, we calculate the most similar words with POS tag vectors using cosine similarity. The results show that most POS tags can be seen as the prototype of words in the category. For instance, the three most similar words to \texttt{NN} is \{\texttt{baby}, \texttt{mother}, \texttt{kitten}\}. The three most similar words to \texttt{VBG} is \{\texttt{takeoff}, \texttt{instrument},  \texttt{wrote}\}. However, there are a few exceptions, such as \texttt{DT} and \texttt{TO}, which return \{\texttt{sahel}, \texttt{pebbly},  \texttt{download}\}, \{\texttt{patent}, \texttt{mickey}, \texttt{ernst}\}, respectively. Regarding the $L_2$ norm, the POS tags with highest values are \{\texttt{NN}, \texttt{PRP\$}, \texttt{TO}, \texttt{PRP}, \texttt{CC}\} and POS tags with lowest values are \{\texttt{WP}, \texttt{POS}, \texttt{EX}, \texttt{VBZ}, \texttt{WRB}\}. Together, the results suggest that only the tags  that satisfy two factors (1) owning higher $L_2$ norm, and (2) sharing similar vector with the corresponding word, such as \texttt{NN}, \texttt{PRP\$} and \texttt{PRP}, achieve higher attention scores. The tags with higher $L_2$ norm but do no share similar vector with the corresponding word, such as \texttt{TO} and \texttt{CC}, get lower attention weights.

\subsection{Relation with Human Attention}
We have built attention models using significant predictors of human reading time, thus we wonder if the proposed  models which are learned from large-scale corpus share similarity with human attention. Therefore, we calculate correlation between our proposed attention models and human reading times in the Dundee corpus \cite{kennedy2003dundee}. 

The Dundee corpus is widely used in researches on human reading and consists of eye-tracking data for 10 subjects reading 20 newswire articles (about 51,000 words). It contains three measures of reading time:
\begin{itemize}
\item	First-pass time (RTfpass) measures the total time spent reading a word before the first fixation on any other word. It is also known as gaze duration.
\item	Go-past time (RTgopast) measures the total time spent reading a word from the first fixation up to the first fixation on a word further to the right. This often includes the reading time on words to the left of the current word.
\item	Right-bounded time (RTrb) measures the total time spent reading a word before the first fixation on a word further to the right. 
\end{itemize}

\begin{table*} \footnotesize
\centering
\caption{Pearson/Spearman rank correlation between human reading times and the attention values predicted by our models. The asterisk means significantly for p-values $<$ 0.0001.}
\begin{tabular}{cccc}
\toprule
Models & RTfpass & RTgopast & RTrb\\	
\midrule
ATT-SUR &	0.412*/0.353* &	0.382*/0.288* & 0.413*/0.349*\\
ATT-POS\tiny{(SCBOW)} &	0.385*/0.307* &	0.370*/0.268* &	0.386*/0.302*\\
ATT-CCG\tiny{(SCBOW)}&	0.430*/0.347* &	0.412*/0.306* &	0.431*/0.342*\\
ATT-POS\tiny{(PP)} &	0.373*/0.123* &	0.353*/0.104* &	0.373*/0.120*\\
ATT-CCG\tiny{(PP)}&	0.390*/0.093* &	0.367*/0.076* &	0.390*/0.091*\\
\bottomrule
\end{tabular}
\end{table*}

Table 3 shows the results of the linear correlation (Pearson rank) and monotonic correlation (Spearman rank) between our models and the averaged human reading times. As it can be seen, the attention weights calculated by our models are highly correlated with human reading time, which suggests that the proposed attention models might have some cognitive plausibility. Furthermore, the attention models with more fine-grained CCG supertags are more relevant to human attention than the attention models with POS tag. Although Surprisal has long been seen as a strong predictor of human reading time, the ATT-CCG model, when incorporated into SCBOW model, reaches even higher correlation scores. These results suggest that our data-driven attention models can predict human reading time to some extent, which may lead to new insights in studies of human reading behaviors.

\section{Related Work}

Learning meaningful sentence representations is the first step towards the goal of language understanding, which has attracted a significant amount of  research attention. Recently, neural network based methods have shown advantage in learning task-specific sentence representations \cite{tai2015improved,chen2015sentence} and general-purpose sentence representations \cite{le2014distributed,kiros2015skip,kenter2016siamese,wieting2016towards}. Our work addresses the problem of \textit{learning general-purpose sentence representations} that capture textual semantics and perform robustly across tasks. 

This paper incorporates the proposed attention model into the averaging  model, which results a weighted additive model. It is necessary to note that the proposed attention model can also be combined with more complex sentence representation models. Though this is beyond the focus of this paper, it remains an interesting direction to explore as a future work. In compositional distributional semantics, the effectiveness of weighted addition is first emphasized by \cite{mitchell2010composition}. They use weighted summation of word vectors in a text, in which weights are parameters (i.e., single value for a word) learned together with word embeddings. This idea is also used in learning word \cite{ling2015not,wang2016cse} and phrase \cite{yu2015learning,wang2017comparison} representations. \cite{ling2015not,wang2016cse} uses weighted summation of context words to predict the target word in learning word representations. The weights are a set of parameters, determining the importance of each word in each relative position in the predefined context window size. Regarding phrase representations, \cite{yu2015learning} employs attention weights as linear combinations of various features, whereas \cite{wang2017comparison} utilizes $L_2$ norm of enhanced word representations as implicit attention weights. Different from these work, we focus on learning representations of sentences and design cognition-inspired methods to calculate attention weights. 

As for sentence representations, \cite{lin2016arxiv,yang2016hierarchical} used hidden layers of the sequence model (i.e., LSTM) as input to a MLP model to get the attention weight of each word. However, this attention mechanism is designed for sequence models and need large supervised training data. In contrast, ours can be easily applied to various kinds of models and can be learned without or with a little supervision. In parallel to our work, \cite{arora2016simple} proposes a similar approach to improve sentence representations by using smooth inverse frequency (SIF) to weight words in a sentence, which achieves comparable results with ours. The most crucial difference is that we employ cognitive plausible factors as auxiliary information to compute attention weights, which brings a lot of scalability, making it applicable to other tasks like recognizing textual entailments. Furthermore, we observe that the performance is improved when applying SIF to ATT-POS (ATT-CCG) model. This indicates that these models capture complementary information and thus an optimal combination of them will probably lead to better attention models, which we leave further investigation to future work.

Another branch of related work is computational models for human reading behaviours \cite{keller2016modeling}. Recently, there are increasing research interests in improving the performance of natural language processing tasks by employing the information in an eye-tracking corpus. For instance, \cite{klerke2016improving} presents a multi-task learning algorithm to improve performance of sentence compression by using an auxiliary training task that is predicting human reading times. \cite{barrett2016weakly} improves the state-of-the-art part of speech tagging (POS) model with various features (i.e., total fixation duration, first pass duration, etc.) derived from an eye-tracking corpus. The work presented in this paper, to our best knowledge, is the first study to improve sentence representations with guidance of human attention in reading. Together, these results indicate that traces or predictors of human cognitive processing, such as the eye-tracking data in reading behaviours, can be used to augment natural language processing models.

\section{Conclusions}
Motivated by the fact that human read sentences by selectively focusing on importance words, in this paper we have demonstrated that introducing such attention mechanism can enhance semantic representations of sentences.

To attach different weights to vectors of the component words in a sentence, we propose two novel approaches that use significant predictors of human reading time to guide the construction of attention models on single sentences. Experimental evaluations show that our approaches lead to substantial gains in accuracy on nearly all 24 SemEval datasets. Qualitative analyses have indicated that the proposed attention models can selectively focus on important words and successfully predict human reading times. These results, coupled with findings from \cite{klerke2016improving} and \cite{barrett2016weakly}, suggest that rich information contained in human cognitive processing can be used to enhance NLP models.

\section*{Acknowledgments}
The research work has been funded by the Natural Science Foundation of China under Grant No. 61333018 and No. 91520204  and supported by the Strategic Priority Research Program of the CAS under Grant No. XDB02070007.

\bibliographystyle{named}
\bibliography{ijcai17}

\begin{thebibliography}{}

\bibitem[\protect\citeauthoryear{Arora \bgroup \em et al.\egroup
  }{2016}]{arora2016simple}
Sanjeev Arora, Yingyu Liang, and Tengyu Ma.
\newblock A simple but tough-to-beat baseline for sentence embeddings.
\newblock {\em arXiv}, 2016.

\bibitem[\protect\citeauthoryear{Barrett \bgroup \em et al.\egroup
  }{2016}]{barrett2016weakly}
Maria Barrett, Joachim Bingel, Frank Keller, and Anders S{\o}gaard.
\newblock Weakly supervised part-of-speech tagging using eye-tracking data.
\newblock {\em The $54^{th}$ Annual Meeting of the Association for
  Computational Linguistics}, page 579, 2016.

\bibitem[\protect\citeauthoryear{Chen \bgroup \em et al.\egroup
  }{2015}]{chen2015sentence}
Xinchi Chen, Xipeng Qiu, Chenxi Zhu, Shiyu Wu, and Xuanjing Huang.
\newblock Sentence modeling with gated recursive neural network.
\newblock {\em Proceedings of EMNLP}, pages 793--798, 2015.

\bibitem[\protect\citeauthoryear{Demberg and Keller}{2008}]{demberg2008data}
Vera Demberg and Frank Keller.
\newblock Data from eye-tracking corpora as evidence for theories of syntactic
  processing complexity.
\newblock {\em Cognition}, 109(2):193--210, 2008.

\bibitem[\protect\citeauthoryear{Ganitkevitch \bgroup \em et al.\egroup
  }{2013}]{ganitkevitch2013ppdb}
Juri Ganitkevitch, Benjamin Van~Durme, and Chris Callison-Burch.
\newblock {PPDB}: The paraphrase database.
\newblock {\em HLT-NAACL}, pages 758--764, 2013.

\bibitem[\protect\citeauthoryear{Hale}{2001}]{hale2001probabilistic}
John Hale.
\newblock A probabilistic earley parser as a psycholinguistic model.
\newblock {\em Proceedings of the NAACL-HLT}, pages 1--8, 2001.

\bibitem[\protect\citeauthoryear{Hill \bgroup \em et al.\egroup
  }{2016a}]{hill2016learning}
Felix Hill, Kyunghyun Cho, and Anna Korhonen.
\newblock Learning distributed representations of sentences from unlabelled
  data.
\newblock {\em Proceedings of NAACL-HLT 2016,}, pages 1367--1377, 2016.

\bibitem[\protect\citeauthoryear{Hill \bgroup \em et al.\egroup
  }{2016b}]{hill2016dict}
Felix Hill, KyungHyun Cho, Anna Korhonen, and Yoshua Bengio.
\newblock Learning to understand phrases by embedding the dictionary.
\newblock {\em Transactions of the Association for Computational Linguistics},
  4:17--30, 2016.

\bibitem[\protect\citeauthoryear{Jozefowicz \bgroup \em et al.\egroup
  }{2016}]{jozefowicz2016exploring}
Rafal Jozefowicz, Oriol Vinyals, Mike Schuster, Noam Shazeer, and Yonghui Wu.
\newblock Exploring the limits of language modeling.
\newblock {\em arXiv preprint arXiv:1602.02410}, 2016.

\bibitem[\protect\citeauthoryear{Keller}{2016}]{keller2016modeling}
Michael Hahn~Frank Keller.
\newblock Modeling human reading with neural attention.
\newblock {\em Proceedings of the Conference on Empirical Methods in Natural
  Language Processing}, 85(95):95, 2016.

\bibitem[\protect\citeauthoryear{Kennedy \bgroup \em et al.\egroup
  }{2003}]{kennedy2003dundee}
Alan Kennedy, Robin Hill, and Jo{\"e}l Pynte.
\newblock The dundee corpus.
\newblock {\em Proceedings of the $12^{th}$ European conference on eye
  movement}, 2003.

\bibitem[\protect\citeauthoryear{Kenter \bgroup \em et al.\egroup
  }{2016}]{kenter2016siamese}
Tom Kenter, Alexey Borisov, and Maarten de~Rijke.
\newblock Siamese cbow: Optimizing word embeddings for sentence
  representations.
\newblock {\em Proceedings of the $54^{th}$ Annual Meeting of the Association
  for Computational Linguistics}, pages 941--951, 2016.

\bibitem[\protect\citeauthoryear{Kiros \bgroup \em et al.\egroup
  }{2015}]{kiros2015skip}
Ryan Kiros, Yukun Zhu, Ruslan~R Salakhutdinov, Richard Zemel, Raquel Urtasun,
  Antonio Torralba, and Sanja Fidler.
\newblock Skip-thought vectors.
\newblock {\em Advances in neural information processing systems}, pages
  3294--3302, 2015.

\bibitem[\protect\citeauthoryear{Klerke \bgroup \em et al.\egroup
  }{2016}]{klerke2016improving}
Sigrid Klerke, Yoav Goldberg, and Anders S{\o}gaard.
\newblock Improving sentence compression by learning to predict gaze.
\newblock {\em Proceedings of NAACL-HLT}, pages 1528--1533, 2016.

\bibitem[\protect\citeauthoryear{Le and Mikolov}{2014}]{le2014distributed}
Quoc~V Le and Tomas Mikolov.
\newblock Distributed representations of sentences and documents.
\newblock {\em Proceedings of the $31^{st}$ International Conference on Machine
  Learning}, pages 1188--1196, 2014.

\bibitem[\protect\citeauthoryear{Levy}{2008}]{levy2008expectation}
Roger Levy.
\newblock Expectation-based syntactic comprehension.
\newblock {\em Cognition}, 106(3):1126--1177, 2008.

\bibitem[\protect\citeauthoryear{Lin \bgroup \em et al.\egroup
  }{2016}]{lin2016arxiv}
Zhouhan Lin, Minwei Feng, Cicero~Nogueira dos Santos, Mo~Yu, Bing Xiang, Bowen
  Zhou, and Yoshua Bengio.
\newblock A self-attentive sentence embedding.
\newblock {\em arXiv}, 2016.

\bibitem[\protect\citeauthoryear{Ling \bgroup \em et al.\egroup
  }{2015}]{ling2015not}
Wang Ling, Lin Chu-Cheng, Yulia Tsvetkov, and Silvio Amir.
\newblock Not all contexts are created equal: Better word representations with
  variable attention.
\newblock {\em Proceedings of the EMNLP}, pages 1367--1372, 2015.

\bibitem[\protect\citeauthoryear{Mikolov \bgroup \em et al.\egroup
  }{2013}]{mikolov2013efficient}
Tomas Mikolov, Kai Chen, Greg Corrado, and Jeffrey Dean.
\newblock Efficient estimation of word representations in vector space.
\newblock {\em arXiv preprint arXiv:1301.3781}, 2013.

\bibitem[\protect\citeauthoryear{Mitchell and
  Lapata}{2010}]{mitchell2010composition}
Jeff Mitchell and Mirella Lapata.
\newblock Composition in distributional models of semantics.
\newblock {\em Cognitive science}, 34(8):1388--1429, 2010.

\bibitem[\protect\citeauthoryear{Rayner}{1998}]{rayner1998eye}
Keith Rayner.
\newblock Eye movements in reading and information processing: 20 years of
  research.
\newblock {\em Psychological bulletin}, 124(3):372, 1998.

\bibitem[\protect\citeauthoryear{Tai \bgroup \em et al.\egroup
  }{2015}]{tai2015improved}
Kai~Sheng Tai, Richard Socher, and Christopher~D Manning.
\newblock Improved semantic representations from tree-structured long
  short-term memory networks.
\newblock {\em Proceedings of the $53^{rd}$ Annual Meeting of the Association
  for Computational Linguistics}, pages 1556--1566, 2015.

\bibitem[\protect\citeauthoryear{Wang and Zong}{2017}]{wang2017comparison}
Shaonan Wang and Chengqing Zong.
\newblock Comparison study on critical components in composition model for
  phrase representation.
\newblock {\em ACM Transactions on Asian and Low-Resource Language Information
  Processing (TALLIP)}, 16(3):16, 2017.

\bibitem[\protect\citeauthoryear{Wang \bgroup \em et al.\egroup
  }{2016}]{wang2016cse}
Yashen Wang, Huang Heyan, Feng Chong, Zhou Qiang, Gu~Jiahui, and Xiong Gao.
\newblock Cse: Conceptual sentence embeddings based on attention model.
\newblock {\em The 54th Annual Meeting of the Association for Computational
  Linguistics}, pages 505--515, 2016.

\bibitem[\protect\citeauthoryear{Wieting \bgroup \em et al.\egroup
  }{2016}]{wieting2016towards}
John Wieting, Mohit Bansal, Kevin Gimpel, and Karen Livescu.
\newblock Towards universal paraphrastic sentence embeddings.
\newblock {\em ICLR}, 2016.

\bibitem[\protect\citeauthoryear{Yang \bgroup \em et al.\egroup
  }{2016}]{yang2016hierarchical}
Zichao Yang, Diyi Yang, Chris Dyer, Xiaodong He, Alex Smola, and Eduard Hovy.
\newblock Hierarchical attention networks for document classification.
\newblock {\em Proceedings of the NAACL}, 2016.

\bibitem[\protect\citeauthoryear{Yu and Dredze}{2015}]{yu2015learning}
Mo~Yu and Mark Dredze.
\newblock Learning composition models for phrase embeddings.
\newblock {\em Transactions of the Association for Computational Linguistics},
  3:227--242, 2015.

\end{thebibliography}

\end{document}